\documentclass[sigconf]{acmart}

\setcopyright{none}
\renewcommand\footnotetextcopyrightpermission[1]{}
\usepackage{microtype}
\usepackage{graphicx}
\usepackage{subcaption}
\usepackage{booktabs}
\usepackage{hyperref}
\usepackage{amsmath}
\usepackage{mathtools}
\usepackage{amsthm}
\usepackage[capitalize,noabbrev]{cleveref}

\theoremstyle{plain}

\theoremstyle{definition}

\theoremstyle{remark}

\AtBeginDocument{%
  }

\begin{document}

\title{Hierarchical Concept-to-Appearance Guidance for Multi-Subject Image Generation}

\author{Yijia Xu}
\authornote{Both authors contributed equally to this research.}
\email{2301213309@stu.pku.edu.cn}
\affiliation{%
  \institution{Peking University}
    \city{Beijing}
    \country{China}
}

\author{Zihao Wang}
\authornotemark[1]
\email{rex.wangzihao@gmail.com}
\affiliation{%
  \institution{Hong Kong University of Science and Technology}
    \city{Hong Kong}
    \country{China}
}

\author{Haokun Gui}
\email{hgui@connect.ust.hk}
\affiliation{%
  \institution{Hong Kong University of Science and Technology}
    \city{Hong Kong}
    \country{China}
}

\author{Jinshi Cui}
\authornote{Corresponding author.}
\email{cjs@cis.pku.edu.cn}
\affiliation{%
  \institution{Peking University}
    \city{Beijing}
    \country{China}
}
\renewcommand{\shortauthors}{Trovato et al.}
\newcommand{\mypara}[1]{\vspace{0.5em}\noindent\textbf{#1}}

\begin{abstract}
  A clear and well-documented \LaTeX\ document is presented as an
  article formatted for publication by ACM in a conference proceedings
  or journal publication. Based on the ``acmart'' document class, this
  article presents and explains many of the common variations, as well
  as many of the formatting elements an author may use in the
  preparation of the documentation of their work.
\end{abstract}



\keywords{Image generation, Multi-Subject Generation, Correspondence Alignment, Cross-Modal Attention}

\begin{abstract}
Multi-subject image generation aims to synthesize images that faithfully preserve the identities of multiple reference subjects while following textual instructions. However, existing methods often suffer from identity inconsistency and limited compositional control, as they rely on diffusion models to implicitly associate text prompts with reference images. In this work, we propose Hierarchical Concept-to-Appearance Guidance (CAG), a framework that provides explicit, structured supervision from high-level concepts to fine-grained appearances. At the conceptual level, we introduce a VAE dropout training strategy that randomly omits reference VAE features, encouraging the model to rely more on robust semantic signals from a Visual Language Model (VLM) and thereby promoting consistent concept-level generation in the absence of complete appearance cues. At the appearance level, we integrate the VLM-derived correspondences into a correspondence-aware masked attention module within the Diffusion Transformer (DiT). This module restricts each text token to attend only to its matched reference regions, ensuring precise attribute binding and reliable multi-subject composition. Extensive experiments demonstrate that our method achieves state-of-the-art performance on the multi-subject image generation, substantially improving prompt following and subject consistency.
\end{abstract}

\maketitle
\fancyhf{}
\fancyhead{}
\fancyfoot[C]{\thepage}
\renewcommand{\headrulewidth}{0pt}
\pagestyle{plain}
\thispagestyle{plain}
\makeatother

\section{Introduction}
\label{sec:intro}

Multi-subject generation aims to synthesize images that follow user-provided editing instructions while faithfully preserving the identities of all referenced subjects. In practical applications such as AI-driven drama production and personalized story illustration, the reference inputs typically comprise multiple characters embedded within complex scenes, while user instructions are often provided as short and concise text. Despite their brevity, such instructions usually imply rich semantic intent, spanning spatial relationships, appearance consistency, and role-specific behaviors. Establishing reliable correspondences between high-level user intent and the fine-grained visual details contained in multiple reference images therefore remains a fundamental challenge.

Recent advances strive to improve subject consistency by refining training objectives. UMO~\cite{UMO} adopts a multi-identity reward learning strategy to enhance identity preservation, whereas MOSAIC~\cite{MOSAIC} enforces cross-image alignment by supervising reference-to-target attention through correspondence masks. However, these approaches mainly regulate the training stage and lack explicit mechanisms that guide the inference process, leading to suboptimal controllability during generation. Moreover, a single DiT~\cite{dit} backbone struggles to handle complex multimodal reasoning when scaling to diverse real-world personalized generation scenarios.

To mitigate the limitations of DiT-only pipelines, several recent works incorporate Vision Language Models. Frameworks such as OmniGen2~\cite{OmniGen2} and Qwen-Image~\cite{Qwen-Image} encode both reference images and user instructions with a Vision Language Models before injecting the extracted multimodal features into a DiT backbone. While these methods benefit from the semantic richness of VLM encodings, they typically treat the Vision Language Model as a passive feature extractor rather than an active reasoning module. Consequently, they fail to fully leverage the VLM’s capacity for semantic grounding and cross-modal alignment.

\begin{figure}[t]
    \centering
    \includegraphics[width=0.8\linewidth]{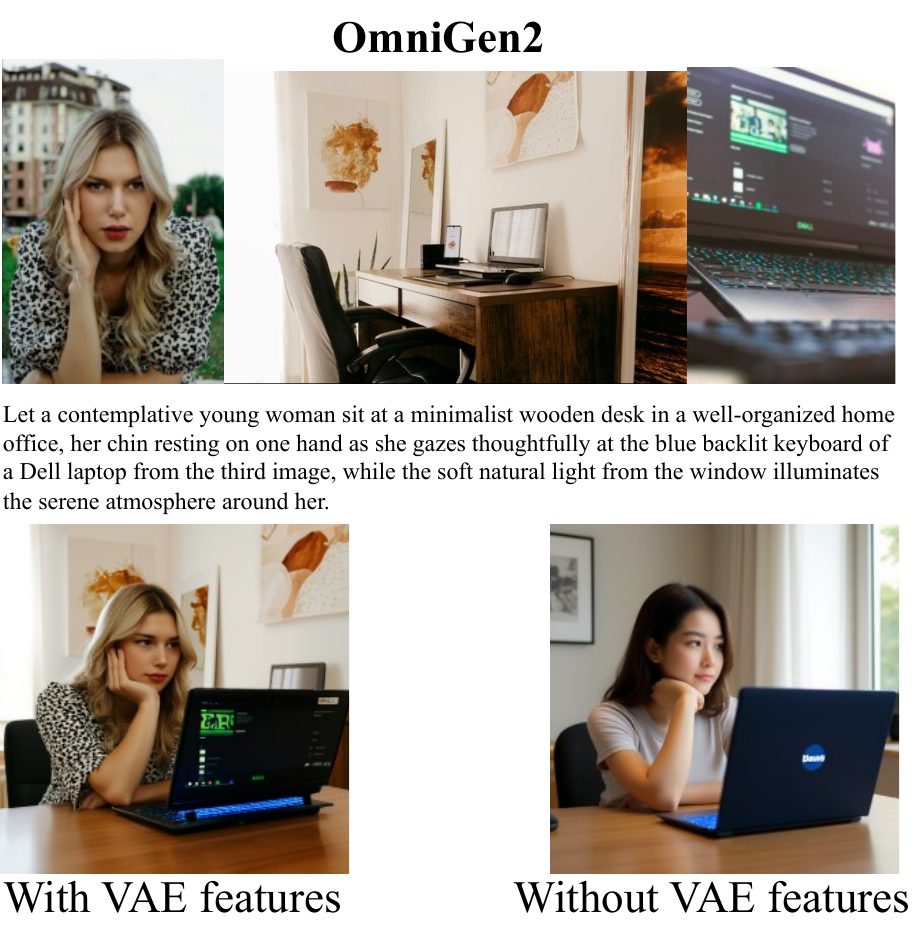}
    \vspace{-0.3cm}
    \caption{Qualitative comparison of OmniGen2~\cite{OmniGen2} under inference with and without using VAE features from reference images.
    }
    \vspace{-0.5cm}
    \label{fig:motivation}
\end{figure}

A core issue persists across existing pipelines: they rely heavily on the DiT backbone to implicitly associate high-level textual instructions with visual cues from multiple reference images. As the number and diversity of references increase, such implicit alignment becomes unreliable. Even when VLM features are included, there remains no principled mechanism to bridge the semantic representations encoded by the VLM with the fine-grained visual features encoded by the VAE. As a result, DiT must infer cross-modal correspondences without explicit structural guidance, leading to unstable identity binding and degraded compositional consistency. Furthermore, while VLM-derived image features inherently contain rich semantic attributes—such as shape, color, and contextual relations—these informative signals are often overshadowed by the model’s over-reliance on low-level VAE features, limiting semantic expressiveness and compositional controllability. 
As shown in \cref{fig:motivation}, removing reference-image VAE features during inference of OmniGen2~\cite{OmniGen2} leads to inconsistencies not only in fine-grained details but also in semantic attributes such as shape and color.
This observation suggests that the model lacks sufficient capability to effectively utilize VLM features.
Effectively integrating the complementary strengths of VLM semantics and VAE fine-grained appearance cues therefore remains an open challenge for multi-subject generation.

To address these limitations, we propose Hierarchical Concept-to-Appearance Guidance (CAG), a multi-subject generation framework that explicitly directs the DiT to perform fine-grained text-to-reference correspondence. CAG leverages a VLM to establish dense word-to-region alignments between editing instructions and reference images, providing explicit semantic grounding. During both training and inference, the DiT is equipped with correspondence-aware masked attention at every layer, constraining each textual token to attend only to the VAE features of its matched reference regions while simultaneously attending to target-image features and VLM-encoded text features. This design filters out irrelevant visual and semantic signals, delivering interpretable and consistent alignment supervision throughout the generative process. Moreover, to further exploit the VLM’s semantic reasoning capabilities, we introduce a VAE dropout strategy that randomly omits reference VAE features during training. By forcing the model to rely solely on VLM-encoded features, CAG strengthens its conceptual grounding and enhances its robustness in scenarios where low-level visual cues are sparse or noisy.

Our main contributions are summarized as follows:
\begin{itemize}
    \item[$\bullet$] We propose \textbf{CAG}, a Hierarchical Concept-to-Appearance Guidance framework for multi-subject image generation, which explicitly incorporates VLM knowledge into diffusion models through concept-level and appearance-level conditioning.

    \item[$\bullet$] At the \emph{conceptual level}, we introduce a VAE dropout training strategy that randomly omits reference VAE features, encouraging the model to rely more strongly on VLM-derived semantic representations and improving its robustness to incomplete low-level visual cues.

    \item[$\bullet$] At the \emph{appearance level}, we employ correspondence-aware masked attention within the DiT backbone, constraining each text token to attend only to its matched reference regions and associated VLM features, thereby enabling precise, localized attribute binding and reliable multi-subject composition.

    \item[$\bullet$] Extensive experiments demonstrate that \textbf{CAG} consistently outperforms existing methods on multi-subject generation tasks involving diverse human identities and scene references, achieving state-of-the-art performance in text fidelity, identity preservation, and compositional quality.
\end{itemize}
\section{Related Work}
\label{sec:related}

\subsection{Multi-subject Driven Image Generation}
Recent advances in image editing models~\cite{MagicBrush,EMU-edit,BAGEL,FLUX-Kontext,InstructP2P,LLMGA,SEED-Data-Edit,Qwen-Image,ominicontrol,xverse,textual_inversion,dreambooth} have shown remarkable progress.
Building upon single-subject generation approaches, a series of multi-subject generation methods have been developed to support reference-guided customization.
UNO~\cite{UNO} introduces a model–data co-evolution strategy to enable consistent generation across multiple references.
DreamOmni2~\cite{DreamOmni2} carefully constructs a three-stage data creation pipeline to acquire high-quality multi-reference training data.
DreamO~\cite{DreamO} computes the loss based on the attention between condition images and the generated image, while MOSAIC~\cite{MOSAIC} introduces two additional losses that respectively utilize reference-to-target latent attention and inter-reference attention to further improve cross-image consistency.
UMO~\cite{UMO} advances this line of research by proposing a multi-identity matching reward that reinforces subject-level consistency.

\subsection{Multimodal Understanding for Reference-Guided Generation}
Multi-subject generation involves both visual and textual inputs, requiring strong multimodal understanding capabilities from the model.
Recent studies have developed unified models~\cite{Chameleon,BAGEL,OneCAT,BLIP3-o,HunyuanImage,Janus-Pro} that integrate both visual understanding and image generation within a single framework.
Some approaches focus on extracting cross-image relationships~\cite{SFNet,Reference-Based,RegionDrag,sift,bay2006surf,dift,sd-dino,geoaware}.
Another line of work~\cite{OmniGen2,Qwen-Image} leverages pretrained vision–language models (VLMs)~\cite{Qwen2.5-VL,Qwen-VL} for their superior multimodal representation ability, feeding both VLM-encoded reference image and text features, along with VAE-encoded visual features, into DiTs.
\section{Methodology}
\label{sec:method}

\begin{figure*}[t]
    \centering
    \includegraphics[width=0.9\linewidth]{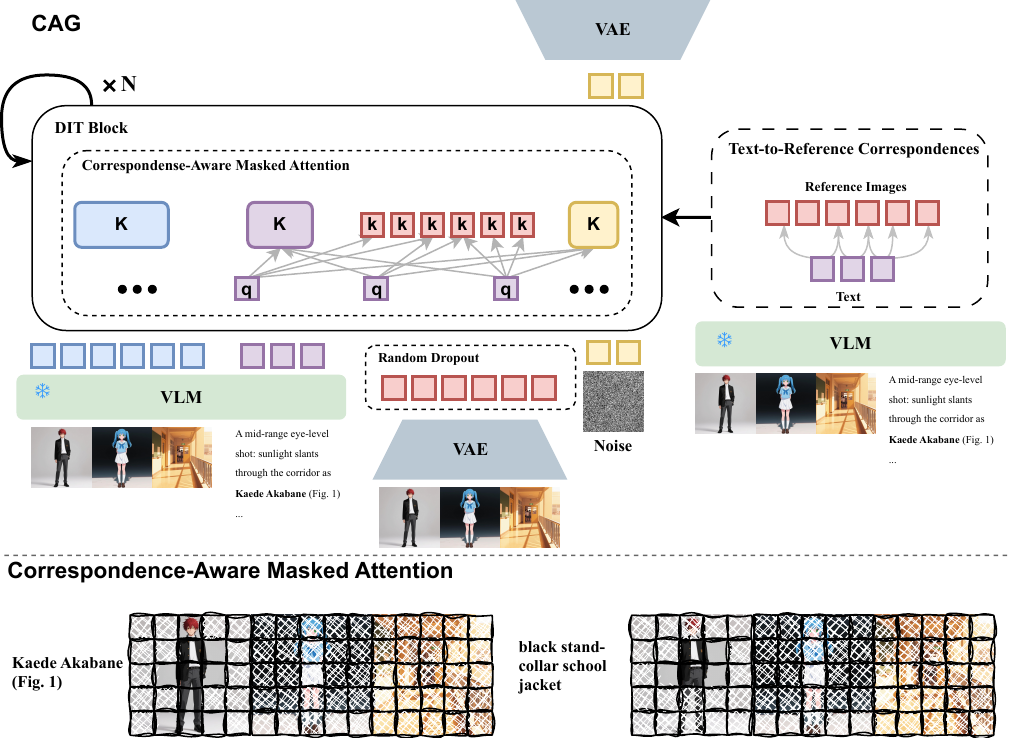}
    \vspace{-0.3cm}
    \caption{\textbf{Overview of CAG.}
    \textbf{Top:} The overall training pipeline of our CAG framework, which integrates the proposed \emph{VAE Dropout} strategy and the \emph{Correspondence-Aware Masked Attention} mechanism. 
    \textbf{Bottom:} Illustration of the proposed Correspondence-Aware Masked Attention, where text tokens are explicitly matched to their associated regions in the reference images, enabling fine-grained and identity-consistent multi-subject generation.
    }
    \vspace{-0.3cm}
    \label{fig:model}
\end{figure*}

\subsection{Preliminary}
\label{pre}
Given a set of $N$ reference images $\{I_i^{r}\}_{i=1}^{N}$ and a textual editing instruction $t^{e}$, the goal of the multi-subject image generation task is to synthesize a target image $I^{t}$ that follows the instruction while preserving and compositing the visual contents from the reference images. Formally, the task can be expressed as
\begin{equation}
I^{t} = \mathcal{F}\big(\{I_i^{r}\}_{i=1}^{N},\, t^{e}\big),
\end{equation}
where $\mathcal{F}(\cdot)$ denotes the conditional image generation process.

In a typical VLM--DiT framework, a VLM is employed to extract multi-modal representations by jointly encoding each reference image and the textual instruction:
\begin{equation}
\mathbf{h}_{i}^{r},\, \mathbf{h}^{e} = \mathrm{VLM}([I_i^{r},\, t^{e}]), \quad i = 1,\dots,N.
\end{equation}
Here, $\mathbf{h}_{i}^{r}$ represents the visual features associated with the $i$-th reference image, and $\mathbf{h}^{e}$ denotes the text-conditioned semantic representation that provides instruction-aware guidance.

Meanwhile, each reference image is also encoded by a VAE encoder to obtain its low-level latent representation:
\begin{equation}
\mathbf{v}_{i}^{r} = \mathrm{VAE_{enc}}(I_i^{r}), \quad i = 1,\dots,N.
\end{equation}

The representations $\{\mathbf{h}_{i}^{r}\}$, $\mathbf{h}^{e}$, and $\{\mathbf{v}_{i}^{r}\}$ serve as conditioning signals for the DiT-based denoising network, which synthesizes the target image $I^{t}$ by integrating semantic guidance from the VLM and fine-grained visual details from the VAE latents.

In the DiT-based architecture, multi-modal features from both the reference and target domains are jointly attended to within each transformer layer.
Specifically, at every layer $\ell$ of the DiT, the query, key, and value representations from the reference-image VLM features, the text VLM features, the target-image VAE features, and the reference-image VAE features are concatenated and processed through a unified self-attention operation.

Formally, let $\mathbf{Q}^{(\ell)}$, $\mathbf{K}^{(\ell)}$, and $\mathbf{V}^{(\ell)}$ denote the query, key, and value matrices in the $\ell$-th attention layer. They are constructed as
\begin{align}
\mathbf{Q}^{(\ell)} = [\,\mathbf{Q}_{\mathrm{VLM}}^{r},\, \mathbf{Q}_{\mathrm{VLM}}^{e},\, \mathbf{Q}_{\mathrm{VAE}}^{t},\, \mathbf{Q}_{\mathrm{VAE}}^{r}\,],
\notag\\
\mathbf{K}^{(\ell)} = [\,\mathbf{K}_{\mathrm{VLM}}^{r},\, \mathbf{K}_{\mathrm{VLM}}^{e},\, \mathbf{K}_{\mathrm{VAE}}^{t},\, \mathbf{K}_{\mathrm{VAE}}^{r}\,],
\notag\\
\mathbf{V}^{(\ell)} = [\,\mathbf{V}_{\mathrm{VLM}}^{r},\, \mathbf{V}_{\mathrm{VLM}}^{e},\, \mathbf{V}_{\mathrm{VAE}}^{t},\, \mathbf{V}_{\mathrm{VAE}}^{r}\,],
\end{align}
where $[\cdot]$ denotes concatenation along the token dimension.

The self-attention for layer $\ell$ is then computed as
\begin{equation}
\mathbf{A}^{(\ell)} = \mathrm{Softmax}\!\left(\frac{\mathbf{Q}^{(\ell)} (\mathbf{K}^{(\ell)})^{\top}}{\sqrt{d}}\right) \mathbf{V}^{(\ell)},
\end{equation}
where $d$ is the feature dimension.  
This unified attention mechanism enables the DiT to perform cross-modal reasoning among four types of representations — reference-image VLM features ($\mathrm{VLM}^{r}$), instruction VLM features ($\mathrm{VLM}^{e}$), reference-image VAE features ($\mathrm{VAE}^{r}$), and target-image VAE features ($\mathrm{VAE}^{t}$) — allowing the network to jointly integrate semantic and visual information at every layer.

\subsection{VAE Dropout Training Strategy}

The image features encoded by the VLM, $\mathbf{h}_{i}^{r}$, contain rich semantic information, analogous to detailed textual descriptions. 
However, during text-conditional pretraining, the DiT has only learned to follow textual guidance and is not trained to directly leverage the VLM-encoded image features. 
DiT, being familiar with the VAE features from pretraining, tends to over-rely on $\mathbf{v}_{i}^{r}$ at the beginning of training, preventing it from effectively exploiting $\mathbf{h}_{i}^{r}$.

To mitigate this issue, we propose a VAE dropout training strategy. 
During training, with a certain probability, the DiT is prevented from using the VAE-encoded reference features $\mathbf{v}_{i}^{r}$ and relies solely on the VLM-encoded textual and image features $\mathbf{h}_{i}^{r}$ to compute the loss. 
Formally, let $p$ denote the dropout probability; for a given training step, the input features to DiT are:
\begin{equation}
\{\mathbf{v}_{i}^{r}, \mathbf{h}_{i}^{r}, \mathbf{h}^{e}\} \quad \longrightarrow \quad
\begin{cases}
\{\mathbf{v}_{i}^{r}, \mathbf{h}_{i}^{r}, \mathbf{h}^{e}\}, & \text{with } 1-p,\\
\{\mathbf{h}_{i}^{r}, \mathbf{h}^{e}\}, & \text{with } p.
\end{cases}
\end{equation}

This forces the DiT to learn to extract detailed semantic information from the VLM-encoded image features $\mathbf{h}_{i}^{r}$, 
thereby enhancing its ability to utilize rich reference information even when textual instructions are concise.

To handle multiple reference images, we introduce a positional offset for the reference-image tokens. 
Let the last token of the $(i-1)$-th reference image have position $(n, m)$. 
Then, the first token of the $i$-th reference image is assigned position $(n+1, m+1)$.
This ensures that the position embeddings of reference-image tokens do not overlap across different references, 
which helps the DiT to distinguish multiple references during both training and inference.

\subsection{Instruction-Reference Alignment}
\label{alignment}

To achieve high-quality multi-subject generation, it is essential to resolve two prerequisite tasks in advance.
First, the model must extract the subset of words in the editing instruction that correspond to the visual contents of the reference images.
Second, for each word, the model must localize the associated visual region within the corresponding reference image, typically represented as a bounding box.

To effectively address these challenges, we leverage the VLM.
For the first task, the reference images and the editing instruction are fed into the VLM, which outputs a set of referential words that correspond to visual subjects present in the reference images:
\begin{equation}
\mathcal{W} = [w_{1}, w_{2}, \dots, w_{M}]
\end{equation}
where each $w_j$ is a noun extracted from the instruction and confirmed to be visually grounded.
For the second task, each referential word $w_j$ is further processed by the VLM to determine its grounding in the reference set. Specifically, for each $w_j$, the VLM outputs the index of the corresponding reference image and its localized bounding box:
\begin{equation}
\bigl(\mathrm{id}_j, [x_{1}^{(j)}, y_{1}^{(j)}, x_{2}^{(j)}, y_{2}^{(j)}]\bigr)
\end{equation}
where $\mathrm{id}_j$ denotes the index of the reference image, and $[x_{1}^{(j)}$, $y_{1}^{(j)}$, $x_{2}^{(j)}$, $y_{2}^{(j)}]$ denotes the bounding box (top-left and bottom-right coordinates).

During both training and inference, the VLM can extract these grounding cues solely based on the user-provided inputs.
These outputs are then used to explicitly guide the attention computations in the DiT, alleviating the need for the DiT to implicitly solve the multimodal grounding task.

\subsection{Correspondence-Aware Masked Attention}
\label{Attention}
In the standard DiT attention layers, each text token from the VLM attends to all other tokens, which implicitly requires the DiT to perform \emph{word--reference region correspondence reasoning}. 
This is challenging and suboptimal for a DiT pretrained on generic image generation tasks, as it is not explicitly trained for such multimodal grounding.

To address this, we leverage the pre-extracted word--bounding box pairs
$\{ (w_j, \mathrm{bbox}_j) \}_{j=1}^{M}$ to \emph{directly guide the DiT attention computation}. 
Specifically, we locate all referential words $w_j$ in the instruction and apply a masked attention that restricts their queries to relevant keys.

Let $\mathbf{Q}_{\mathrm{VLM}}^{w_j}$ denote the query corresponding to word $w_j$, 
$\mathbf{K}_{\mathrm{VAE}}^{r, \mathrm{bbox}_j}$ denote the keys of the reference-image VAE tokens inside bounding box $\mathrm{bbox}_j$, 
and $\mathbf{K}_{\mathrm{VLM}}^{e}, \mathbf{K}_{\mathrm{VAE}}^{t}$ denote the keys of the text VLM tokens and the target-image VAE tokens, respectively.
The attention for word $w_j$ is then computed as:
\begin{align}
\mathbf{A}_{w_j} &= 
\mathrm{Softmax}\Bigg(
\frac{\mathbf{Q}_{\mathrm{VLM}}^{w_j} 
\big[ \mathbf{K}_{\mathrm{VLM}}^{e},\, \mathbf{K}_{\mathrm{VAE}}^{t},\, \mathbf{K}_{\mathrm{VAE}}^{r, \mathrm{bbox}_j} \big]^\top}
{\sqrt{d}}
\Bigg) \notag\\
&\quad \times
\big[ \mathbf{V}_{\mathrm{VLM}}^{e},\, \mathbf{V}_{\mathrm{VAE}}^{t},\, \mathbf{V}_{\mathrm{VAE}}^{r, \mathrm{bbox}_j} \big],
\end{align}
where all irrelevant reference-image VAE tokens outside $\mathrm{bbox}_j$ and all reference-image VLM tokens are masked out.

This \emph{Correspondence-Aware Masked Attention} mechanism allows the DiT to focus only on relevant multimodal associations, explicitly injecting the word--region correspondence into the attention process, while avoiding the need for DiT to implicitly solve the challenging grounding task.
The correspondence between referential words and reference regions is explicitly encoded in the attention mechanism. 
As a result, when performing attention, the target-image VAE tokens can easily attend to the correct reference-image regions while following the guidance of the text tokens. 

Formally, this establishes a binding between the textual instruction and the relevant reference regions, allowing the DiT to naturally extend its capability of following the text to also follow the reference image content. 
This mechanism enables the model to achieve both high text fidelity and high reference-image consistency in the generated results.

\section{Experiments}
\label{sec:exp}

\begin{figure*}[t]
    \centering
    \includegraphics[width=0.9\linewidth]{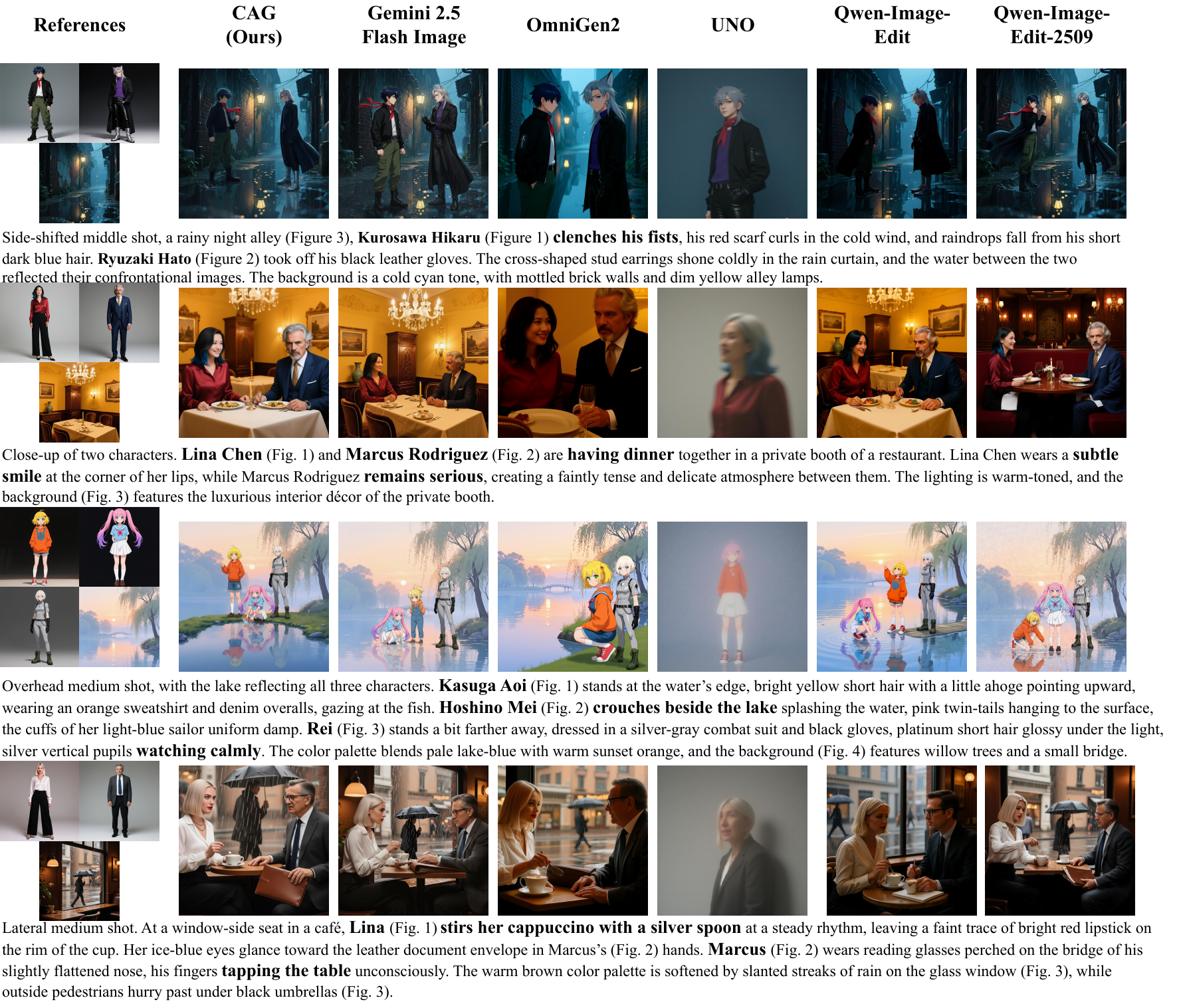}
    \vspace{-0.3cm}
    \caption{Qualitative comparison with different methods on multi-subject driven image generation.
    }
    \vspace{-0.5cm}
    \label{fig:visual}
\end{figure*}

\subsection{Implementation Details}
\label{imp}
We build our method upon Qwen-Image-Edit~\cite{Qwen-Image}, a recent VLM-DiT framework designed for single-image editing. 
Since the original model only supports single-reference editing, we extend it to the multi-subject generation setting.
During training, the VLM is frozen, while the DiT module is fully fine-tuned on a multi-subject dataset containing approximately 24k examples with diverse character and scene references.

We employ the AdamW~\cite{adamw} optimizer with a learning rate of 1e-5, training for 9k steps using a total batch size of 8.
Following our design in Sec.~\ref{sec:method}, we apply a \textbf{VAE dropout} with a probability of 0.5.
Additional ablation results on the VAE dropout probability are reported in the Appendix.
To support classifier-free guidance, we adopt a text dropout with a probability of 0.1.
During inference, the number of denoising steps is set to 25, and the classifier-free guidance scale is set to 4.0.
We utilize Qwen3-VL-30B-A3B-Instruct~\cite{qwen3-vl} to extract \textit{word–bounding box} pairs for both training and inference.

\subsection{Evaluation Metrics}
\label{eval}
The test set contains 300 samples, each consisting of multiple reference subjects and one reference scene. 
Following OmniGen2~\cite{OmniGen2} and VIEScore~\cite{ku2023viescore}, we employ the multi-modal model GPT-4.1~\cite{gpt4-1} to evaluate the quality of multi-subject image generation results. 
GPT-4.1 assesses each generated image along two dimensions: Prompt Following (PF) and Subject Consistency (SC), with scores ranging from 0 to 10. 
An Overall Score is further computed as the geometric mean of PF and SC, providing a balanced measure of both instruction adherence and subject fidelity.
Following UNO~\cite{UNO}, we additionally report DINO, CLIP-I, and CLIP-T scores for completeness. 
DINO and CLIP-I measure subject similarity using cosine similarity between image-level features. 
However, these metrics provide limited explainability and are primarily designed for simple single-reference settings, making them inadequate for complex multi-subject generation. 
CLIP-T computes the cosine similarity between the prompt and the image CLIP embeddings, which is effective for short image captions but does not adequately capture the semantics of complex image editing instructions. 
Therefore, we mainly rely on \textbf{PF, SC, and Overall as the primary evaluation metrics} for comparison, while DINO, CLIP-I, and CLIP-T are reported for reference.

\subsection{Main Results}
\label{main-result}
\begin{table*}
  \centering
  \resizebox{0.9\textwidth}{!}{
  \begin{tabular}{l c c c c c c}
    \toprule
    \textbf{Method} & \textbf{PF}$\uparrow$ & \textbf{SC}$\uparrow$ & \textbf{Overall}$\uparrow$ & DINO$\uparrow$ & CLIP-I$\uparrow$ & CLIP-T$\uparrow$ \\
    \midrule 
    Gemini 2.5 Flash Image \cite{banana} & 6.903 & 7.819 & 7.308 & 0.511 & 0.708 & 0.318 \\
    OmniGen2 \cite{OmniGen2} & 4.644 & 6.044 & 5.242 & 0.504 & 0.702 & 0.315 \\
    UNO \cite{UNO} & 1.344 & 3.097 & 1.685 & 0.503 & 0.704 & 0.245 \\
    Qwen-Image-Edit \cite{Qwen-Image} & 5.599 & 6.719 & 6.074 & 0.500 & 0.701 & 0.312 \\
    Qwen-Image-Edit-2509 \cite{Qwen-Image} & 5.890 & 6.613 & 6.178 & 0.496 & 0.701 & 0.317 \\
    CAG (Ours) & \textbf{7.308} & \textbf{7.906} & \textbf{7.568} & \textbf{0.526} & \textbf{0.720} & \textbf{0.319} \\
    \bottomrule
  \end{tabular}
  }
  \caption{\textbf{Comparison on the multi-subject image generation task.}
  Each test sample contains multiple reference subjects and one reference scene.
  ``PF'' denotes Prompt Following. ``SC'' denotes Subject Consistency.
  Since Qwen-Image-Edit~\cite{Qwen-Image} only supports single-image input, all reference images are concatenated into a single composite image to serve as input during inference. 
  }
  \vspace{-0.5cm}
  \label{tab:compare}
\end{table*}

\mypara{Quantitative Results.}
\Cref{tab:compare} reports the quantitative results on the multi-subject image generation task. 
Our method achieves the highest scores across all metrics, demonstrating superior ability in both instruction adherence and subject identity preservation. 
Specifically, our model attains 7.308 in Prompt Following and 7.906 in Subject Consistency, leading to an Overall Score of 7.568. 
These results demonstrate that our approach establishes a new SOTA for multi-subject image generation.

When compared with OmniGen2~\cite{OmniGen2} and Qwen-Image-Edit-2509~\cite{Qwen-Image}, which share the VLM-DiT architectural design, our method achieves clear advantages.
Specifically, our PF, SC, and Overall scores surpass OmniGen2 by +2.664, +1.862, and +2.326, respectively, and exceed Qwen-Image-Edit-2509 by +1.418, +1.293, and +1.390. 
Relative to our base model Qwen-Image-Edit~\cite{Qwen-Image}, our method improves PF by +1.709, SC by +1.187, and Overall Score by +1.494. 
This demonstrates the high quality of our generated results.

\mypara{Qualitative Results.}
\Cref{fig:visual} presents the qualitative comparison results on the multi-subject image generation task.
Our method demonstrates the best overall performance in both text alignment and reference consistency.

For instance, in the second row of \cref{fig:visual}, Gemini 2.5 Flash Image, OmniGen2, and Qwen-Image-Edit exhibit noticeable discrepancies in the character’s hair color compared to the reference image. 
Qwen-Image-Edit-2509 produces inaccuracies in facial details and background consistency. 
By comparison, CAG better preserves fine-grained character-specific attributes while maintaining coherence with the reference background.

\subsection{Ablation Studies}
\label{subsec:ablation}

\begin{table}
  \centering
  \resizebox{\linewidth}{!}{
  \begin{tabular}{cc|ccc}
    \toprule
    Mask Attn. & VAE Dropout & PF$\uparrow$ & SC$\uparrow$ & Overall$\uparrow$ \\
    \midrule
     & & 6.797 & 7.653 & 7.177 \\
    $\checkmark$ & & 7.063 & 7.853 & 7.415 \\
     & $\checkmark$ & 7.177 & 7.712 & 7.402 \\
    $\checkmark$ & $\checkmark$ & 7.308 & 7.906 & 7.568 \\
    \bottomrule
  \end{tabular}
  }
  \caption{\textbf{Effectiveness of proposed methods.}
  ``Mask Attn.'' denotes correspondence-aware masked attention.
  ``VAE Dropout'' denotes the VAE dropout training strategy.
  ``PF'' denotes Prompt Following. ``SC'' denotes Subject Consistency.
  }
  \vspace{-0.5cm}
  \label{tab:ablation}
\end{table}

\begin{table}
  \centering
  \resizebox{0.9\linewidth}{!}{
  \begin{tabular}{c|ccc}
    \toprule
    Method & PF$\uparrow$ & SC$\uparrow$ & Overall$\uparrow$ \\
    \midrule
    w/o VAE droput & 6.591 & 5.807 & 6.146 \\
    w/ VAE dropout & 7.258 & 7.354 & 7.273 \\
    \bottomrule
  \end{tabular}
  }
  \caption{Quantitative comparison under inference \textbf{without using VAE features from reference images}.
  \textbf{w/o VAE dropout} denotes the model trained without the VAE dropout strategy and evaluated without reference-image VAE features, while \textbf{w/ VAE dropout} denotes the model trained with the VAE dropout strategy under the same inference setting.
  }
  \vspace{-0.5cm}
  \label{tab:vaedrop}
\end{table}

\begin{figure}[t]
    \centering
    \includegraphics[width=\linewidth]{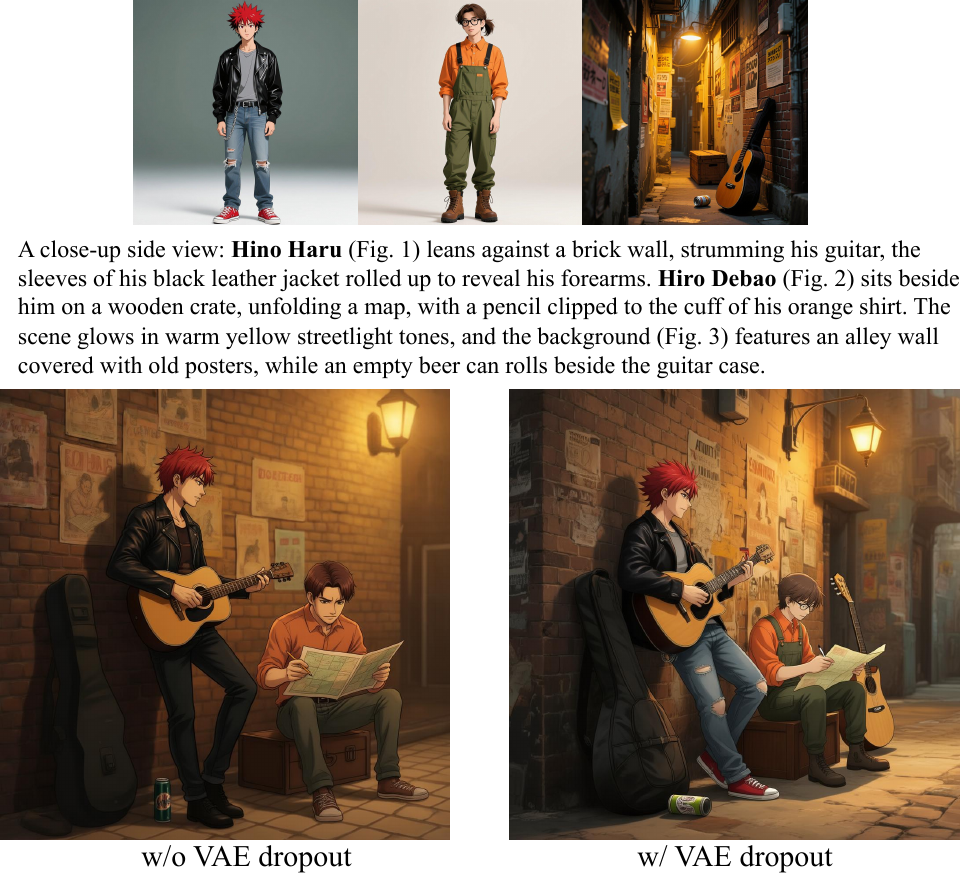}
    \vspace{-0.3cm}
    \caption{Qualitative comparison under inference \textbf{without using VAE features from reference images}.
  \textbf{w/o VAE dropout} denotes the model trained without the VAE dropout strategy and evaluated without reference-image VAE features, while \textbf{w/ VAE dropout} denotes the model trained with the VAE dropout strategy under the same inference setting.
    }
    \vspace{-0.5cm}
    \label{fig:vaedrop_fig}
\end{figure}

\mypara{Effectiveness of Proposed Methods.}
To verify the effectiveness of our proposed methods, we perform ablation experiments on the correspondence-aware masked attention and the VAE dropout training strategy, as shown in \cref{tab:ablation}.

When only the correspondence-aware masked attention is applied, the model achieves a PF score of 7.063, an SC score of 7.853, and an Overall score of 7.415. 
Compared with the baseline (6.797 / 7.653 / 7.177), this corresponds to improvements of +0.266, +0.200, and +0.238, respectively. 
By explicitly guiding each attention layer in DiT with the correspondence between textual tokens and reference image regions, the model effectively filters out irrelevant visual information, leading to better prompt adherence and subject consistency. 

When only the VAE dropout strategy is used, the model achieves 7.177 in PF, 7.712 in SC, and 7.402 in Overall, outperforming the baseline by +0.380, +0.059, and +0.225, respectively. 
This shows that occasionally removing VAE features during training encourages DiT to depend more on VLM-derived semantic information, allowing it to capture details that are not explicitly mentioned in the text but are present in the reference images. 

When both the correspondence-aware masked attention and the VAE dropout strategy are jointly applied, the model achieves the best overall performance with a PF score of 7.308, an SC score of 7.906, and an Overall score of 7.568. 
Their combination enables the model to achieve significant improvements in both prompt adherence and reference consistency.

As shown in \cref{tab:vaedrop}, when VAE features are not used during inference, 
the model trained with the VAE dropout strategy achieves a PF score of 7.258 and an SC score of 7.354, resulting in an Overall score of 7.273. 
Compared to the model trained without VAE dropout (Overall 6.146), this represents a significant improvement of +1.127 in overall performance. 

As shown in \cref{fig:vaedrop_fig}, when VAE features are not used during inference, 
the model trained without VAE dropout (left) produces incorrect details such as mismatched clothing and missing glasses.
In contrast, the model trained with VAE dropout (right) generates results that accurately reproduce the reference clothing.

\begin{figure}[t]
    \centering
    \includegraphics[width=\linewidth]{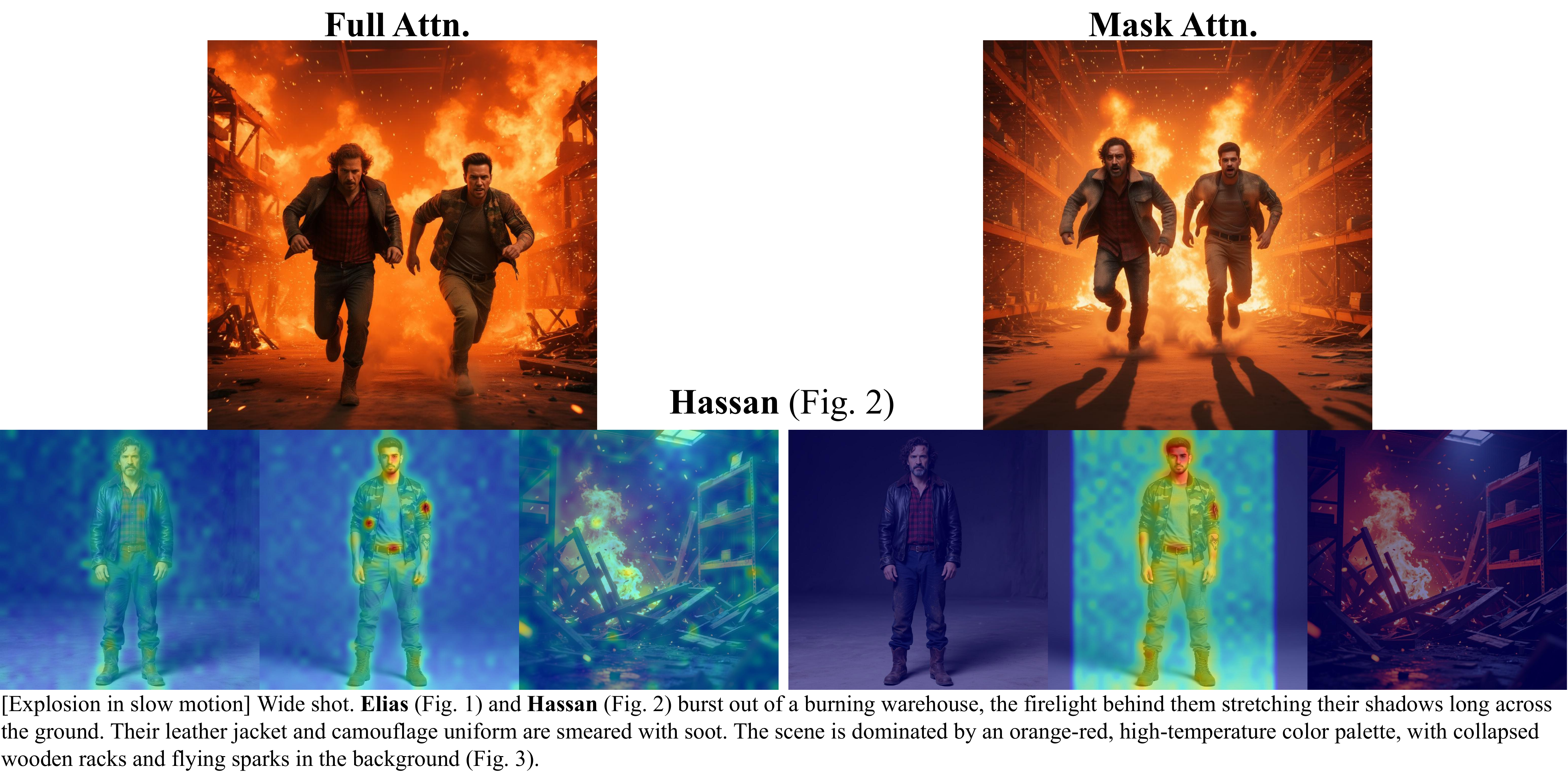}
    \vspace{-0.3cm}
    \caption{Visualization of attention maps from textual phrases to the VAE features of the reference images.
    ``Full Attn.'' denotes using full attention over all tokens, while ``Mask Attn.'' denotes our proposed correspondence-aware masked attention.
    The example visualizes the attention map for the phrase ``Hassan (Fig.~2)''. 
    }
    \vspace{-0.5cm}
    \label{fig:attention}
\end{figure}

\mypara{Visualization of Text-to-Reference Attention Maps.}
As shown in \cref{fig:attention}, the attention maps between textual phrases and the VAE features of the reference images demonstrate that our proposed correspondence-aware masked attention enables the textual descriptions to accurately focus on their corresponding visual references.

In the example, when using Full Attn., the phrase ``Hassan (Fig.~2)'' exhibits dispersed attention over all reference images. 
Although part of the attention falls on the second reference image, the highlighted region is small and fails to cover the entire face or the overall clothing of the character.
In contrast, with Mask Attn., the attention of ``Hassan (Fig.~2)'' is sharply concentrated on the correct reference subject, particularly on the facial region, leading to noticeably improved subject consistency in the generated results.

\begin{table}[t]
\centering
\resizebox{\columnwidth}{!}{
\begin{tabular}{lccccc}
\toprule
Method & Steps & sec/image$\downarrow$ & PF$\uparrow$ & SC$\uparrow$ & Overall$\uparrow$ \\
\midrule
OmniGen2 & 25 & 94.66 & 4.644 & 6.044 & 5.242 \\
Qwen-Image-Edit-2509 & 40 & 290.46 & 5.890 & 6.613 & 6.178 \\
Qwen-Image-Edit-2509 & 25 & 183.75 & -- & -- & -- \\
CAG (Ours) & 25 & \textbf{92.93} & \textbf{7.308} & \textbf{7.906} & \textbf{7.568} \\
\bottomrule
\end{tabular}
}
\caption{Comparison of sampling speed.}
\vspace{-0.5cm}
\label{tab:sampling_speed}
\end{table}

\mypara{Ablation on Speeds.}
As shown in \cref{tab:sampling_speed}, despite incorporating additional components, our CAG achieves the highest PF, SC, and Overall scores while also delivering the fastest generation speed. Compared with Qwen-Image-Edit-2509, CAG not only outperforms its 40-step setting in quality but also runs substantially faster than its 25-step setting.

This efficiency gain mainly stems from our masked-attention design in the DiT architecture, which substantially reduces the computational cost of attention. Although our method introduces an additional VLM-based bounding-box extraction step, the overall pipeline remains highly efficient, enabling both superior generation quality and faster inference.

\begin{table}[t]
\centering
\label{tab:vlm_features}
\resizebox{\columnwidth}{!}{
\begin{tabular}{ccc|ccc}
\toprule
VLM image features & VAE Dropout & Mask Attn. & PF$\uparrow$ & SC$\uparrow$ & Overall$\uparrow$ \\
\midrule
                  &              & \checkmark & 7.053 & 7.855 & 7.408 \\
\checkmark        &              & \checkmark & 7.063 & 7.853 & 7.415 \\
\checkmark        & \checkmark   & \checkmark & 7.308 & 7.906 & \textbf{7.568} \\
\bottomrule
\end{tabular}
}
\caption{Effectiveness of VLM features.}
\vspace{-0.5cm}
\label{tab:vlm_features}
\end{table}

\mypara{Ablation on Effectiveness of the VLM.}
To clarify the source of representational strength, we further ablate the use of reference-image VLM features during both training and inference, as shown in \cref{tab:vlm_features}. The results suggest that VLM features alone provide only marginal gains without VAE Dropout, improving the Overall score from 7.408 to 7.415. In contrast, when VAE Dropout is enabled, VLM features become substantially more effective, increasing the Overall score to 7.568 and improving all metrics. This indicates that VLM and VAE features are complementary: VLM features enhance conceptual alignment, while VAE features preserve appearance fidelity, and our VAE Dropout strategy better activates their joint benefit.
\section{Conclusion}
\label{sec:conclusion}
In this work, we present \textbf{CAG}, a framework that couples high-level VLM semantics with low-level appearance cues. 
CAG prevents the DiT from being overloaded with multimodal understanding and can be readily incorporated into unified multimodal understanding–generation models.

Extensive experiments show that CAG achieves state-of-the-art performance in prompt following and subject consistency. Our findings highlight the value of combining conceptual and appearance-level guidance: high-level semantics provide stable, interpretable grounding, while low-level appearance cues preserve fine-grained subject details. This hierarchical synergy offers a promising direction for controllable and semantically grounded image generation in future unified architectures.

\bibliographystyle{ACM-Reference-Format}
\bibliography{sample-base}

\newpage
\begin{figure*}[t]
    \centering
    \includegraphics[width=\linewidth]{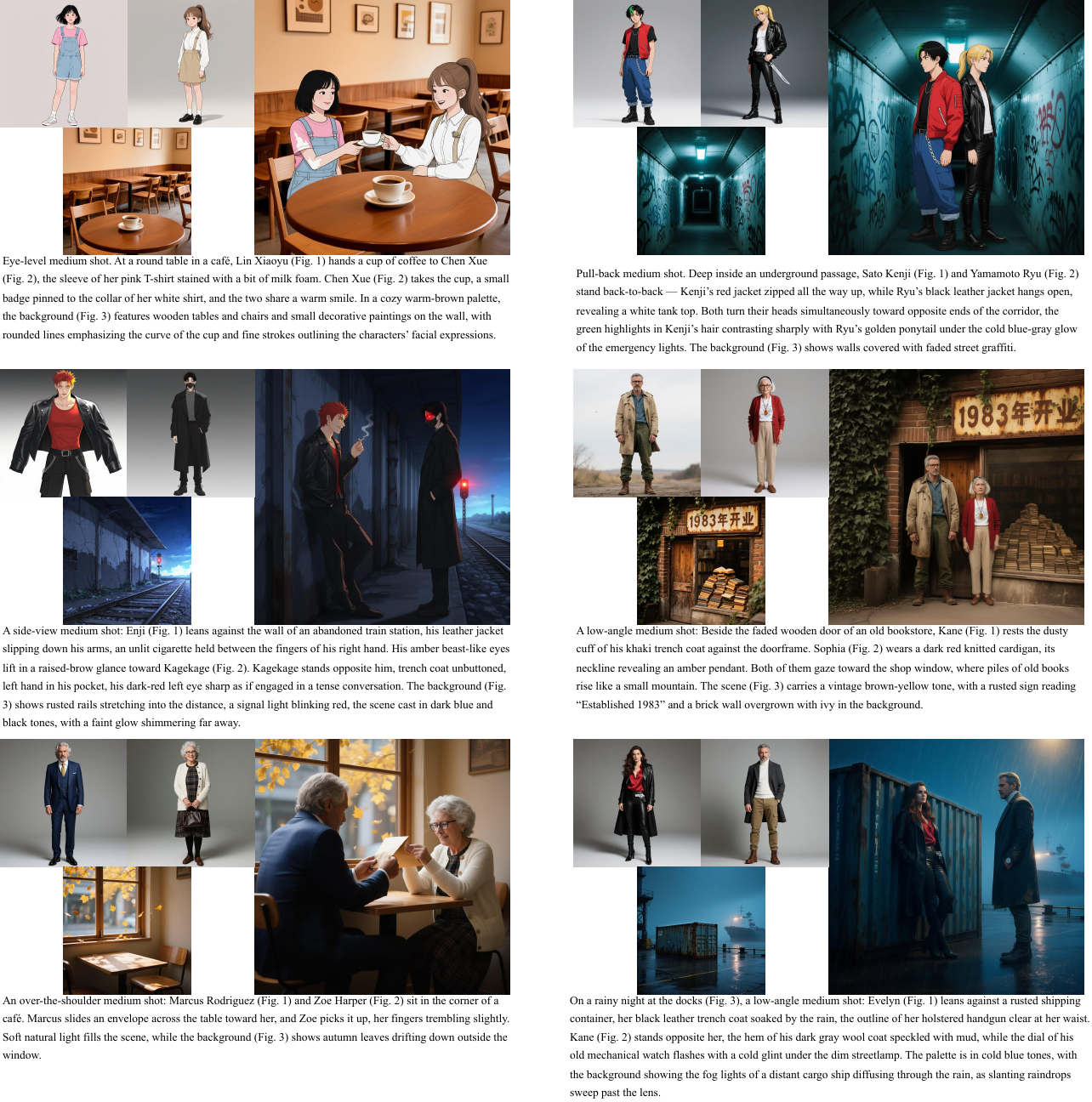}
    \caption{Generated samples from CAG.}
    \label{fig:more1}
\end{figure*}

\begin{figure*}[t]
    \centering
    \includegraphics[width=\linewidth]{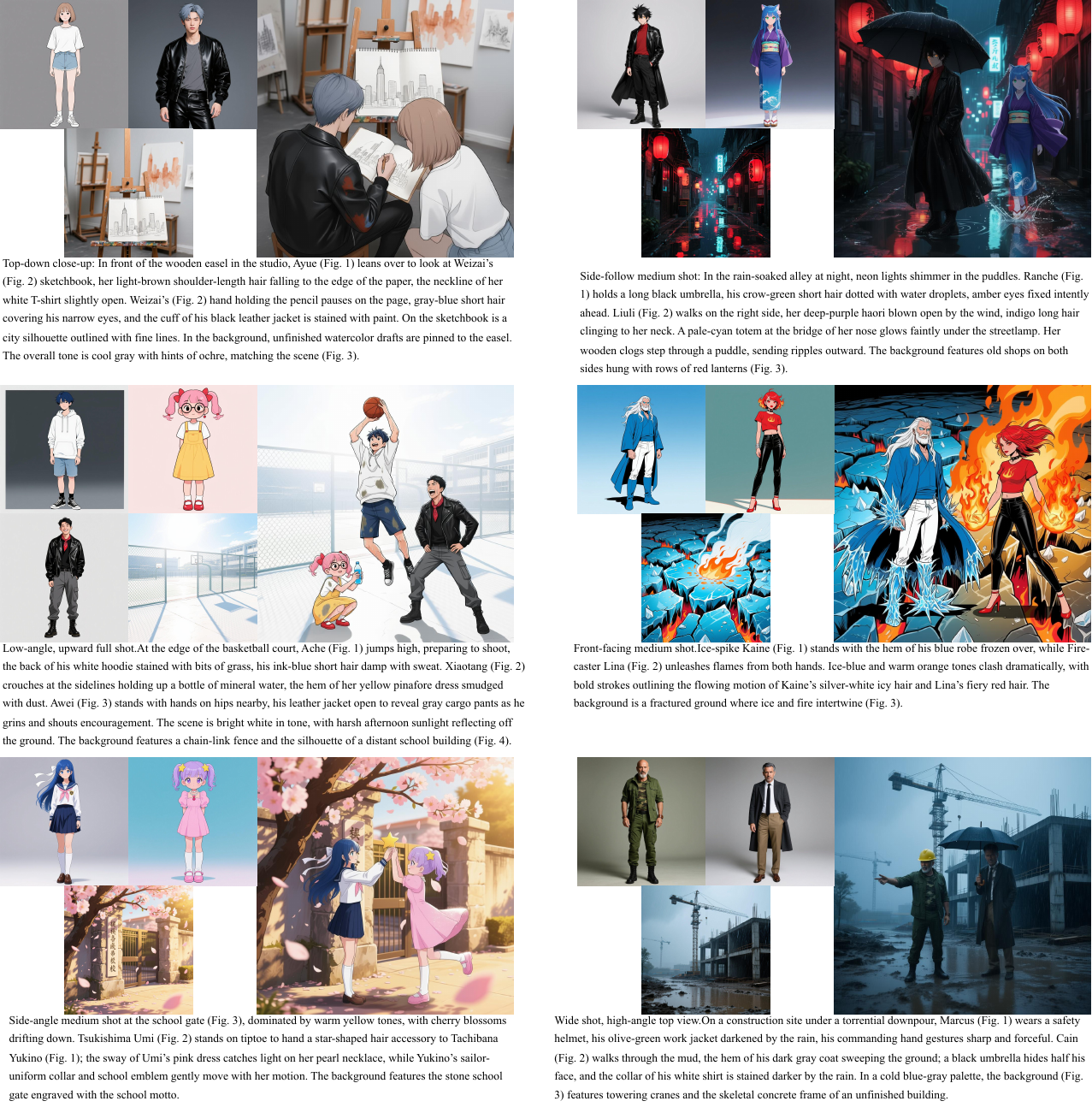}
    \caption{Generated samples from CAG.}
    \label{fig:more3}
\end{figure*}

\clearpage
\appendix

\setcounter{tocdepth}{-1}
\tableofcontents

\section{Details of Instruction-Reference Alignment}
\label{sec:prompt}

\begin{figure*}[!t]
    \centering
    \begin{subfigure}[t]{0.32\textwidth}
        \centering
        \includegraphics[width=\linewidth]{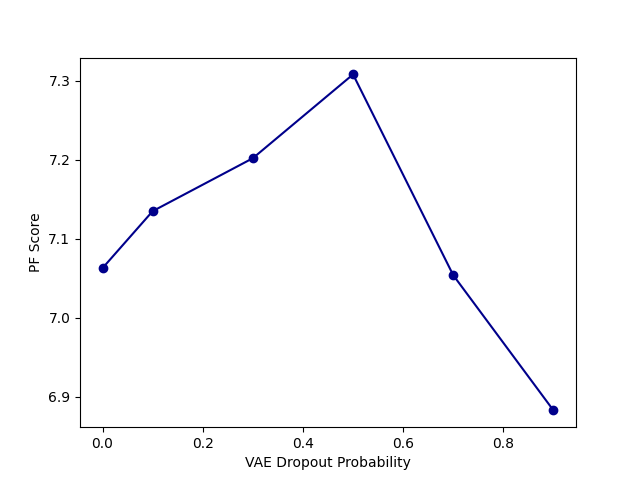}
        \caption{PF score.}
        \label{fig:PF}
    \end{subfigure}
    \hfill
    \begin{subfigure}[t]{0.32\textwidth}
        \centering
        \includegraphics[width=\linewidth]{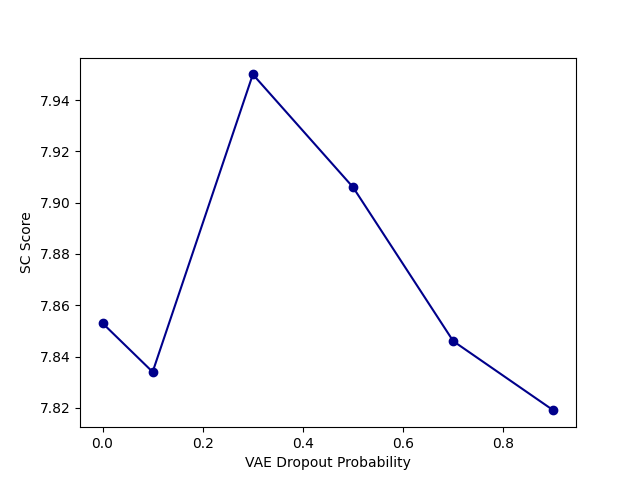}
        \caption{SC score.}
        \label{fig:SC}
    \end{subfigure}
    \hfill
    \begin{subfigure}[t]{0.32\textwidth}
        \centering
        \includegraphics[width=\linewidth]{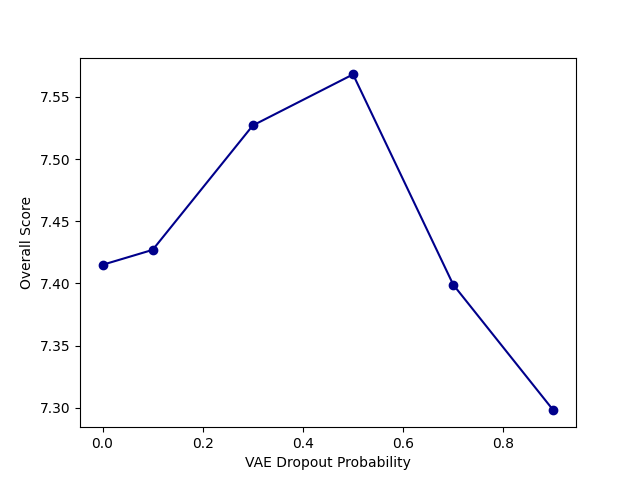}
        \caption{Overall score.}
        \label{fig:overall}
    \end{subfigure}
    \caption{\textbf{Effect of VAE dropout probability.}
    We report PF, SC, and Overall scores under different VAE dropout probabilities.
    Moderate dropout improves performance, while excessive dropout degrades all metrics.}
    \label{fig:drop_prob}
\end{figure*}

Instruction-reference alignment decomposes the grounding process into two subtasks:
(i) identifying words in the editing instruction that are semantically grounded in the reference images, and
(ii) localizing the corresponding visual regions for each identified word.
We use Qwen3-VL-30B-A3B-Instruct~\cite{qwen3-vl} for both stages.

\begin{figure*}[!t]
    \centering
    \begin{subfigure}[t]{0.485\textwidth}
        \centering
        \includegraphics[width=\linewidth]{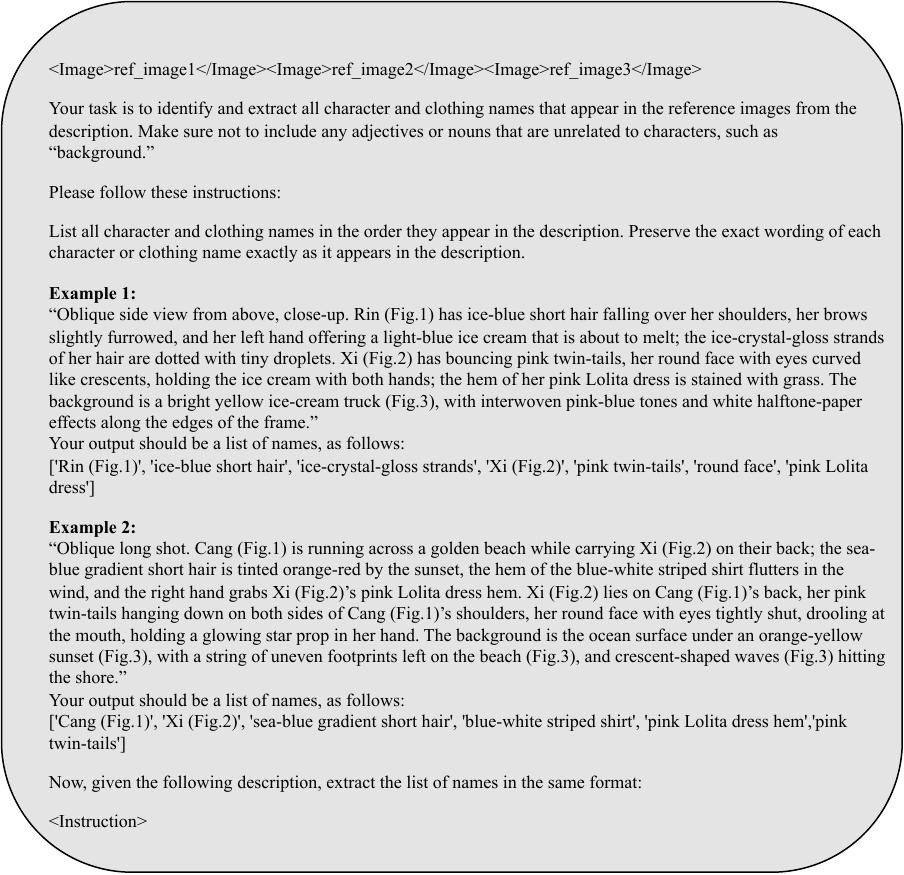}
        \caption{Prompt for extracting reference-related words.}
        \label{fig:obj_prompt}
    \end{subfigure}
    \hfill
    \begin{subfigure}[t]{0.485\textwidth}
        \centering
        \includegraphics[width=\linewidth]{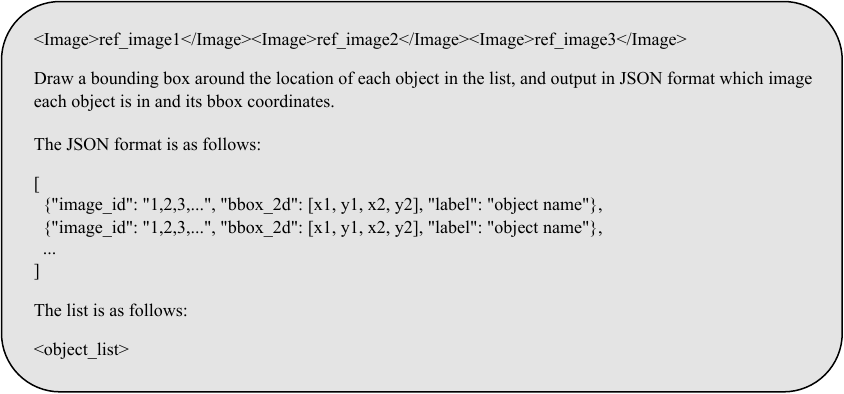}
        \caption{Prompt for predicting word-level bounding boxes.}
        \label{fig:bbox_prompt}
    \end{subfigure}
    \caption{\textbf{Prompts used for instruction-reference alignment.}
    The first prompt identifies words in the instruction that are grounded in the reference images, and the second prompt localizes the corresponding visual regions.}
    \label{fig:alignment_prompts}
\end{figure*}

\begin{figure*}[!t]
    \centering
    \includegraphics[width=\textwidth]{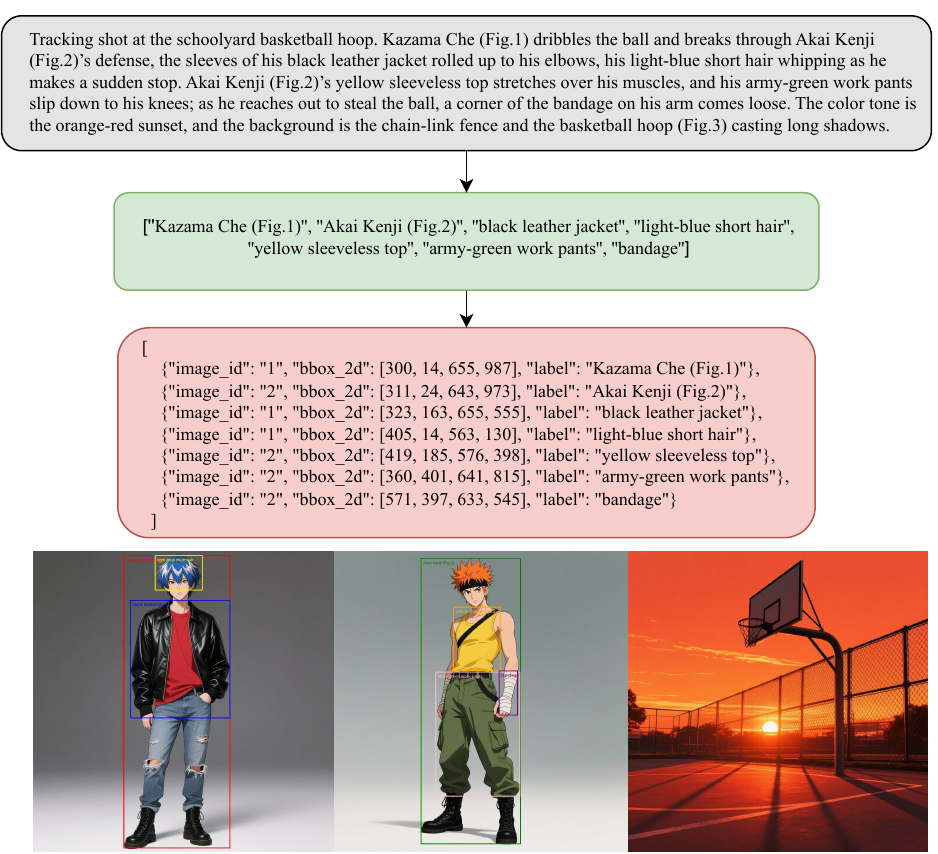}
    \caption{\textbf{Example of instruction-reference alignment.}
    Given an editing instruction and reference images, the VLM extracts reference-grounded words and predicts their corresponding regions.
    The bottom visualizations show the localized regions on the reference images.}
    \label{fig:alignment_case}
\end{figure*}

\Cref{fig:alignment_prompts} summarizes the two prompts used in our pipeline.
The first prompt extracts reference-related words from the editing instruction, while the second prompt predicts word-level bounding boxes for the extracted terms.
As illustrated in \cref{fig:alignment_case}, the VLM first parses the editing instruction to identify nouns associated with the reference images and then localizes their corresponding visual regions.
The visualizations at the bottom of \cref{fig:alignment_case} show the grounded regions on the reference images.

\section{Ablation on VAE Dropout Probability}
\label{sec:drop}

With Correspondence-Aware Masked Attention enabled, we train our model using different VAE dropout probabilities.
As shown in \cref{fig:drop_prob}, PF, SC, and Overall scores~\cite{OmniGen2} follow a consistent pattern: moderate dropout produces the best results across all metrics.
Performance improves as the dropout probability increases from $0$ to the middle range, peaking around $0.3$--$0.5$.
PF and Overall reach their highest values at a dropout probability of $0.5$, while SC peaks at $0.3$.

Randomly dropping VAE features encourages the model to rely more strongly on the fine-grained semantic cues encoded in VLM features.
However, excessive dropout weakens the model's ability to exploit detailed VAE-based appearance information.
An appropriate dropout probability therefore balances semantic guidance and appearance preservation, yielding consistent gains in both prompt following and subject consistency.

\section{More Results}
\label{sec:result}

\begin{figure*}[!t]
    \centering
    \includegraphics[width=\textwidth]{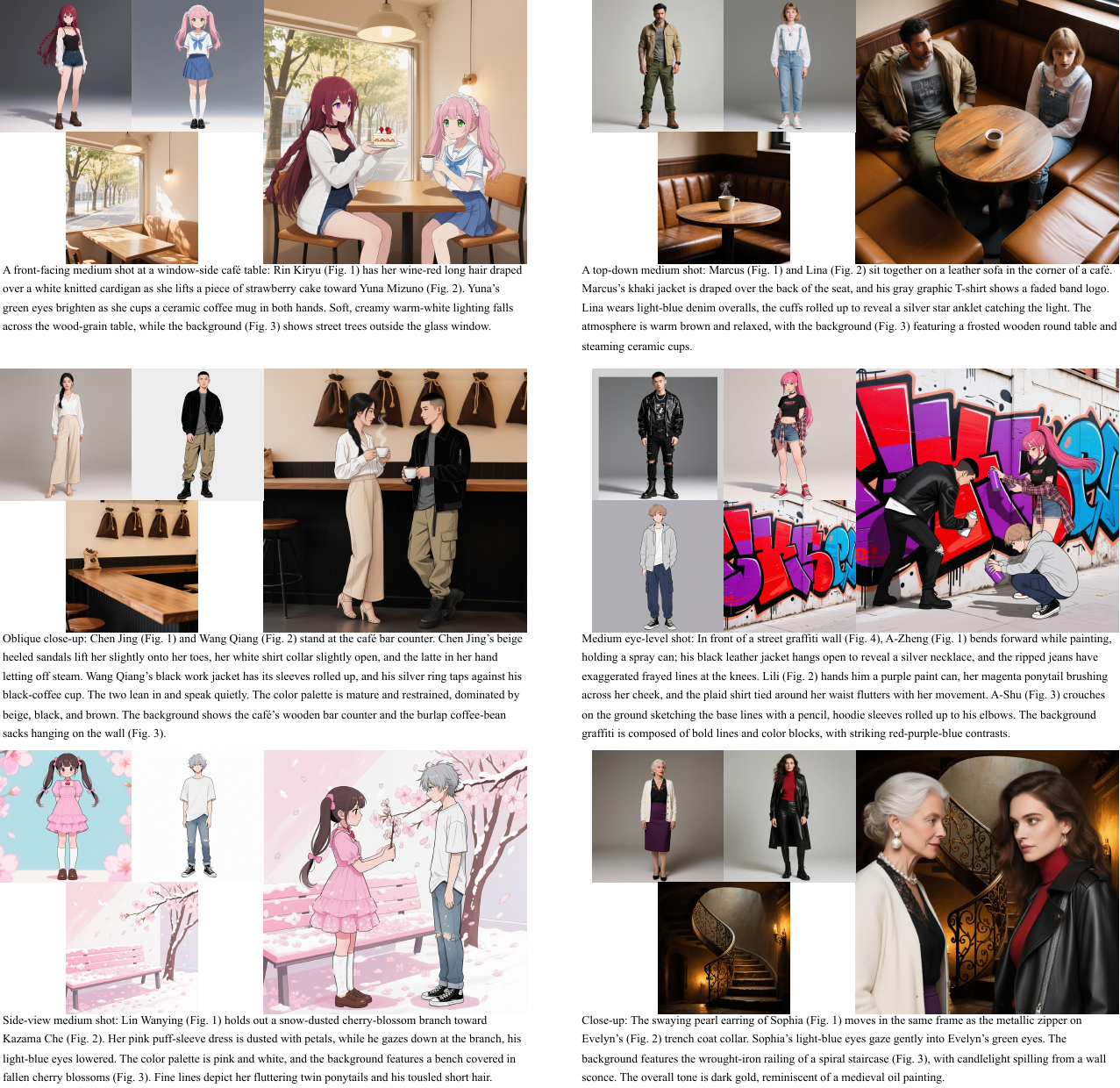}
    \caption{\textbf{Additional qualitative results generated by CAG.}
    CAG follows the editing instructions, arranges multiple referenced characters coherently, and preserves fine-grained subject appearance.}
    \label{fig:more_results}
\end{figure*}

We present additional qualitative results generated by CAG in \cref{fig:more_results}.
The results show that CAG faithfully follows the action descriptions specified in the editing instructions and places multiple characters coherently within the scene defined by the reference images.
CAG also adapts the global composition according to viewpoint directives in the instruction, demonstrating strong controllability over scene layout.
The generated images preserve fine-grained subject appearance, confirming the effectiveness of our VAE dropout training strategy and the proposed Correspondence-Aware Masked Attention.

\begin{table}[!t]
    \centering
    \caption{\textbf{Comparison on the multi-subject image generation task.}
    CAG outperforms Qwen-Image-Edit-2511 across all metrics.}
    \label{tab:multi_subject_comparison}
    \setlength{\tabcolsep}{8pt}
    \renewcommand{\arraystretch}{1.12}
    \resizebox{0.82\linewidth}{!}{
    \begin{tabular}{lccc}
        \toprule
        \textbf{Method}
        & \textbf{PF}$\uparrow$
        & \textbf{SC}$\uparrow$
        & \textbf{Overall}$\uparrow$ \\
        \midrule
        Qwen-Image-Edit-2511
        & 6.035
        & 6.707
        & 6.304 \\
        CAG (Ours)
        & \textbf{7.308}
        & \textbf{7.906}
        & \textbf{7.568} \\
        \bottomrule
    \end{tabular}
    }
\end{table}

\end{document}